\documentclass{article}
\usepackage{arxiv}

\usepackage{graphicx} % Required for inserting images
\usepackage{dsfont}
\usepackage{enumitem}
\usepackage{pgfplots}
\usepackage{subcaption}
\usepackage{booktabs}
\usepackage{float}
\usepackage{xspace}
\usepackage{multirow}
\usepackage{balance}
\PassOptionsToPackage{sort&compress}{natbib}
\usepackage{natbib}
\usepackage{url}
\usepackage{hyperref}
\usepackage{amsmath}
\usepackage[]{mdframed}
\usepackage{alltt} 

\pgfplotsset{compat=newest}

\newcommand{\Comment}[1]{}
\newcommand{\datamorgana}{DataMorgana}

\newcommand{\yoelle}[1]{\noindent{\textcolor{blue}{\{{\bf Yoelle:} {\em #1}\}}}}
\newcommand{\david}[1]{\noindent{\textcolor{red}{\{{\bf David:} {\em #1}\}}}}
\newcommand{\simo}[1]{\noindent{\textcolor{magenta}{\{{\bf Simone:} {\em #1}\}}}}
\newcommand{\guy}[1]{\noindent{\textcolor{orange}{\{{\bf Guy:} {\em #1}\}}}}
\newcommand{\oren}[1]{\noindent{\textcolor{cyan}{\{{\bf Oren:} {\em #1}\}}}}
\newcommand{\ran}[1]{\noindent{\textcolor{teal}{\{{\bf Ran:} {\em #1}\}}}}
\Comment{
\newcommand{\yoelle}[1]{}
\newcommand{\david}[1]{}
\newcommand{\simo}[1]{}
\newcommand{\guy}[1]{}
\newcommand{\oren}[1]{}
\newcommand{\ran}[1]{}
}

\begin{document}

\title{SIGIR 2025 -- LiveRAG Challenge Report}
%\oren{What about: SIGIR'2025 LiveRAG Challenge Report}

\author{
David Carmel, Simone Filice, Guy Horowitz, Yoelle Maarek, Oren Somekh, Ran Tavory\\
Technology Innovation Institute (TII), Haifa, Israel\\
{\bf Mehdi Ghissassi}\textsuperscript{$\dagger$}, {\bf Edo Liberty}\textsuperscript{$\ddagger$}{\bf Roy Miara}\textsuperscript{$\ddagger$}\\
\textsuperscript{$\dagger$}AI71, Abu Dhabi, UAE \textsuperscript{$\ddagger$} Pinecone, New York, USA}

\maketitle

\begin{abstract}
The LiveRAG Challenge at SIGIR 2025, held between March and May 2025, provided a competitive platform for advancing Retrieval-Augmented Generation (RAG) technologies. Participants from academia and industry were invited to develop a RAG-based question-answering system using a fixed corpus (Fineweb-10BT) and a common open-source LLM (Falcon3-10B-Instruct). The goal was to facilitate challenging comparisons of retrieval and prompting strategies. During the Live Challenge Day, 70 teams from 27 different countries provided answers and supportive information to 500 unseen questions within a strict two-hour time window. Evaluation was conducted in two stages: first an automated LLM-as-a-judge approach was used to compute correctness and faithfulness score, then a manual review of top ranked submissions was conducted. The finalists were announced on June 12, 2025, with prizes awarded during the LiveRAG Workshop at SIGIR 2025 in Padua, Italy.
\end{abstract}

\section{Overview}
Retrieval-Augmented Generation (RAG) has emerged as a widely accepted methodology for enhancing the effectiveness of large language models (LLMs), particularly for question-answering tasks \citep{lewis2020retrieval,izacard2022few, Gao+al:24a}. Given a user request, a RAG system searches auxiliary sources to augment the request with relevant content~\citep{lewis2020retrieval}. RAG is attracting significant attention from the AI and IR communities, yet quality assessment of RAG systems is still an open challenge \citep{pradeep2025great,es2024ragas}.

The goal of the LiveRAG Challenge\footnote{\url{https://liverag.tii.ae/}} was to allow research teams, across academia and industry, to advance their RAG research by evaluating their question answering solution and comparing the performance of their system with other teams, on a fixed external corpus (derived from the publicly available Fineweb\footnote{\url{https://huggingface.co/datasets/HuggingFaceFW/fineweb}}), and a fixed open-source LLM ( Falcon3-10B-Instruct\footnote{\url{https://huggingface.co/tiiuae/Falcon3-10B-Instruct}}), during a strict two-hour time window.

Participants were expected to apply their own approach for key elements of the RAG system, such as query rewrite, text retrieval, prompt generation, etc., and integrate their solution with Falcon3-10B-Instruct for answer generation. 
Additionally, participant teams were given early access to TII’s \datamorgana{}~\citep{filice2025generatingdiverseqabenchmarks}, a synthetic Q\&A generation tool, helping them generate benchmarks for training and testing their RAG systems. The shared corpus and the fixed generation model put the focus on the retrieval aspect of RAG, ensuring fair comparison of retrieval and prompting strategies. 

LiveRAG differs from similar competitions such as CRAG \citep{yang2024crag} and the RAG Track in TREC 2024 \citep{thakur2025support} in two key aspects. First, LiveRAG is ``pseudo-live'', hence its name, in order to mitigate the risk of over-tuning to the test set. Namely, participants had to submit their answers to a large set of unseen questions under a strict two-hour time limit. Second, in order to include participants who might not have access to expensive computational resources, selected teams were awarded \$1500 Amazon AWS\footnote{\url{https://aws.amazon.com/}} compute credits, as well as \$750 Pinecone\footnote{\url{https://www.pinecone.io/}} credits (graciously provided by our sponsors, AWS and Pinecone) for building their RAG solution. These resources, in addition to pre-built indices and complimentary access to the AI71 platform to run \datamorgana{} and Falcon3, significantly lowered the entry barrier to the Challenge, making LiveRAG both rigorous and inclusive.

Following recent studies that highlight the limitation of LLM-as-a-judge approach to assess quality, \cite{rahmani2025,soboroff2025don}
%\yoelle{add another citation from Ian Soboroff maybe that predates this one, David?} 
we decided to apply a two-stage evaluation methodology. In  first stage, the correctness and faithfulness scores (See \S{\ref{sec:eval}}) of each submission were automatically computed via a carefully prompted strong LLM (different from the one used by the participants). Then manual assessment of more than half of the leading submissions was conducted by independent annotators.
The finalists presented their results at the LiveRAG workshop at the SIGIR'2025 conference, during which winners were announced and prizes were awarded: \$5000 for the winner, \$3000 for the second, and \$2000 for the third ranked system.
%\oren{if we publish this report after the workshop we should include the winners details}
%\david{This is not common in such reports.}

\section{Challenge's Resources}
%Fixing the external corpus and the LLM for answer generation enabled participants to focus on the retrieval component and on evaluating their RAG solution,  while providing fair comparison between teams. 
In the following, we describe the main resources that were made available to participants during the challenge period.

\subsection{External Content Source}
The Fineweb dataset \citep{penedo2024the} consists of cleaned and de-duplicated Web content from CommonCrawl\footnote{\url{https://commoncrawl.org}}.
%\oren{consider "While Fineweb is relatively cleaner than other web-scale datasets, it still contains some toxic or offensive material and non-English pages, which increases the difficulty of the Challenge."} 
While Fineweb is relatively cleaner than other web-scale datasets, it still contains some toxic or offensive material and non-English pages, which increases the difficulty of the challenge.
For the LiveRAG challenge, the RAG external repository was fixed to Fineweb-10BT\footnote{\url{https://huggingface.co/datasets/HuggingFaceFW/fineweb/viewer/sample-10BT}}, a randomly sampled subset of 15M web documents from Fineweb. 
Although additional content sources were permitted, none of the teams chose to do so, as all evaluation questions were guaranteed to be answerable by Fineweb-10BT alone.

\subsection{Pre-built retrieval indices}
Participants had the option to build their own search indices over Fineweb-10BT, or to use prebuilt sparse and dense indices. 
To prepare these indices, Fineweb-10BT documents were segmented into non-overlapping,
sentence-based chunks of up to 512 tokens each, using the LlamaIndex sentence splitter\footnote{\url{https://www.llamaindex.ai/}}. After segmenting each document into sentences, consecutive sentences were aggregated, preserving sentence boundaries, until the chunk length limit was reached.
\begin{itemize}
\item {\bf OpenSearch Sparse index:} The resulting chunks were indexed into a BM25-based Sparse index implemented on the OpenSearch platform\footnote{\url{https://opensearch.org}}, with OpenSearch default parameter settings.
\item {\bf Pinecone Dense index:}
% Each chunk was embedded into a 768-dimensional vector using the E5-base-v2 model \citep{wang2022text}. The embedding vectors were then indexed by Pinecone, which enables efficient nearest-neighbor search over them. \yoelle{Roy/Edo to add some more details?}
%\oren{please have a look and check if the citation for E5-base-v2 is appropriate} 
Each chunk was embedded into a 768-dimensional vector using the E5-base-v2 model \cite{wang2024textembeddingsweaklysupervisedcontrastive}. The embedding vectors were indexed using Pinecone’s Slab architecture \cite{ingber2025accurate}. Under this architecture, every set of vectors is partitioned into non-overlapping, immutable components called ``Slabs". Data is first written to a Level-0 Slab, and compacted incrementally as new data is inserted. In practice, data is indexed in different ways depending on the Slab level and index size. Smaller corpora are indexed using a random projection algorithm \cite{ailon.2009}, while larger corpora are indexed using IVF-PQ \cite{jegou.2011}, and HNSW \cite{malkov.2020}. Although these latter algorithms are slower at indexing time, they provide significantly faster retrieval, especially for larger datasets. For the LiveRAG Challenge, this trade-off made them the preferred solution given the size of the corpus and the strict response time constraints.
\Comment{Each chunk was embedded into a 768-dimensional vector using the E5-base-v2 model\oren{add a proper citation} [24]. The embedding vectors were indexed using Pinecone’s Slab architecture \cite{ingber2025accurate}. In this architecture, every set of vectors is partitioned into non-overlapping, immutable\oren{consider "components"} parts called ``slabs".\oren{"Data is first written to a Level..."} First, data is written into Level-0 slab,\oren{"and compacted incrementally as new data is inserted."} then compacted as new data is inserted. In practice, data is indexed in\oren{"different ways depending..."} multiple ways, depending on the slab level and index size. Smaller corpora are indexed\oren{omit "after"} after using the random projection algorithm \cite{ailon.2009}, while larger\oren{"corpora"} ones are indexed using IVF-PQ \cite{jegou.2011},\oren{use "and" instead "as well as"} as well as HNSW \cite{malkov.2020}.\oren{"Although these latter algorithms are slower at indexing time, they provide significantly faster retrieval..."} The latter algorithms are slower than random projection at indexing time, but allow much faster retrieval, especially for larger datasets.\oren{"For the LiveRAG Challenge, this trade-off made them the preferred solution given the size of the corpus and the strict response time constraints."} This was thus the solution of choice for the LiveRAG Challenge due to its large corpus and strict response time constraints.}
\end{itemize}

\subsection{Answer Generator}
One key design decision of the LiveRAG Challenge was to require all competitors to use the same LLM for answer generation, namely Falcon3-10B-Instruct,
an instruction-tuned language model developed at the Technology Innovation Institute\footnote{\url{https://www.tii.ae/}} (TII), which, at the time of writing, achieves a state-of-the-art performance among LLMs of similar size. It represents a significant advancement in the Falcon family, trained on two Tera-tokens of diverse datasets and fine-tuned on 1.2 million samples of specialized content.

Participating teams received free access to Falcon3-10B-Instruct through the AI71 platform. For other tasks in the RAG pipeline, besides answer generation, teams were allowed to use other LLMs, and computational tools in general, provided they do not exceed 10B parameters, to align with the challenge theme of using only moderately sizes resources for building a RAG system.

\subsection{\datamorgana{}: Synthetic Benchmark Generator}
Participants were granted early access to \datamorgana{} \citep{filice2025generatingdiverseqabenchmarks,filice2025dmacl},
a novel tool that allows RAG developers to generate synthetic benchmarks from a given corpus via configuration instructions. Configuring the expected question and answer types, and the personas of the hypothetical users posing them, allows to enhance the benchmark’s diversity, realism, and quality.

Participants were given AI71 credits to freely use \datamorgana{} for training and evaluating their systems before the Live Challenge Day. \datamorgana{} was also used for generating an unseen test set of Q\&A pairs, based on Fineweb-10BT, intended for RAG systems' evaluation during the live event (See \S{\ref{sec:eval}}). The test set was intentionally generated to provide diverse test cases of varying complexity for both retrieval and answer-generation stages.
The \datamorgana{} configuration used for test set generation  comprises seven distinct question categorizations, 
%with each question assigned one category from each categorization, 
including answer-type categorization (e.g., factoid, yes/no, list, comparison, multi-aspect, etc.), question-formulation categorization (e.g., natural question vs. search query), linguistic-correctness categorization (i.e., varying levels of spelling or grammar errors), answer-control categorization (concise vs. comprehensive answers), and more.
Notably, questions labeled as comparison or multi-aspect type require at least two documents from the corpus for answering.
Additionally, each question was generated with one user persona out of four options: novice, expert, researcher, and journalist. Table \ref{tab:qa} presents several example Q\&A pairs generated by \datamorgana{} based on randomly selected documents from Fineweb. 

%\yoelle{I am okay sharing the configuration settings but I am commenting out all the filtering information - too much details on DataMorgana.}
% To ensure the high-quality of the generated Q\&A pairs, we applied two LLM-based filtering steps:
% %\simo{Shall we say that these filtering stages are LLM-based?}:
% \begin{itemize}
%  \item Document selection and filtering (pre-generation): We first retrieved documents related to a curated set of topics, then filtered them based on factors such as factuality, credibility, and toxicity. 
%  %For the Challenge we used a filtered subset of\oren{not sure we need to specify how many docs were used. It's too low} $\sim2K$ documents for Q\&A generation.
% \item Q\&A filtering (post-generation): This step ensures that the generated Q\&As meet specific quality criteria --- the question should be understandable without reading the source document, should not be time-sensitive, and can be answered by the source document.
%\end{itemize}

\begin{table}
\scriptsize
\caption{\label{tab:qa} Examples of Q\&As generated by \datamorgana{} for the Live Challenge Day.}
\begin{tabular}{p{0.28\textwidth}p{0.7\textwidth}}
Question & Answer\\
\toprule
Is barefoot outdoor activity linked to reduced inflammation? &
Yes, earthing/grounding has been demonstrated to decrease pro-inflammatory markers in minutes. It is recommended that patients play bare-footed in their yards for at least 15 to 20 minutes a day. This practice also increases energy and can decrease jet lag.\\
\midrule
% How deep can fish survive in the ocean trenches? &
% Fish can survive up to 8,100 meters deep. Beyond this depth, no fish live, possibly due to their inability to cope with the highest pressures. \simo{Let's put a different example. Here the answer might be imprecise: \url{https://www.nhm.ac.uk/discover/news/2023/april/deepest-ever-fish-filmed-depth-8336-metres.html}}\\
What happened to DJ Khaled when he showed his credit card in an online video?	&
DJ Khaled was robbed of around \$80,000 after he flashed his credit card during a 'motivational' video he posted online. Viewers of the video could clearly see his card details and stole from him within moments of the video being posted. \\
\midrule
Which solar cycles were identified in the Qilian mountains temperature study? &
This study identified centennial cycles of 113.6-169 years and 500 years that were possibly associated with solar variations. However, this claim is considered weak due to statistical issues and the non-stationarity in the temperature reconstruction during the 20th Century.\\
\midrule
What are the main symptoms of fungal toenail infections? &
The main symptoms of fungal toenail infections include noticeably thicker toenails, white or yellowish-brown discoloration on toenails, dry, brittle, or ragged toenails, foul smell coming from toes, and unusual toenail shape.\\
\bottomrule
\end{tabular}
\end{table}

\section{Participants}

Seventy teams from 27 countries submitted applications to participate in the LiveRAG challenge. After a thorough review of the submitted applications, while considering factors such as novelty, feasibility, and clarity, we allocated compute resources to 40 teams, including overall more than 140 members from 16 countries, with approximately 77\% of them affiliated with academic institutions.
The distribution of accepted teams by country is presented in Table \ref{tab: accepted teams distribution}.% Figure~\ref{fig:country-pie}.

\Comment{
\begin{figure}[htb]
\centering
\includegraphics[width=0.4\textwidth]{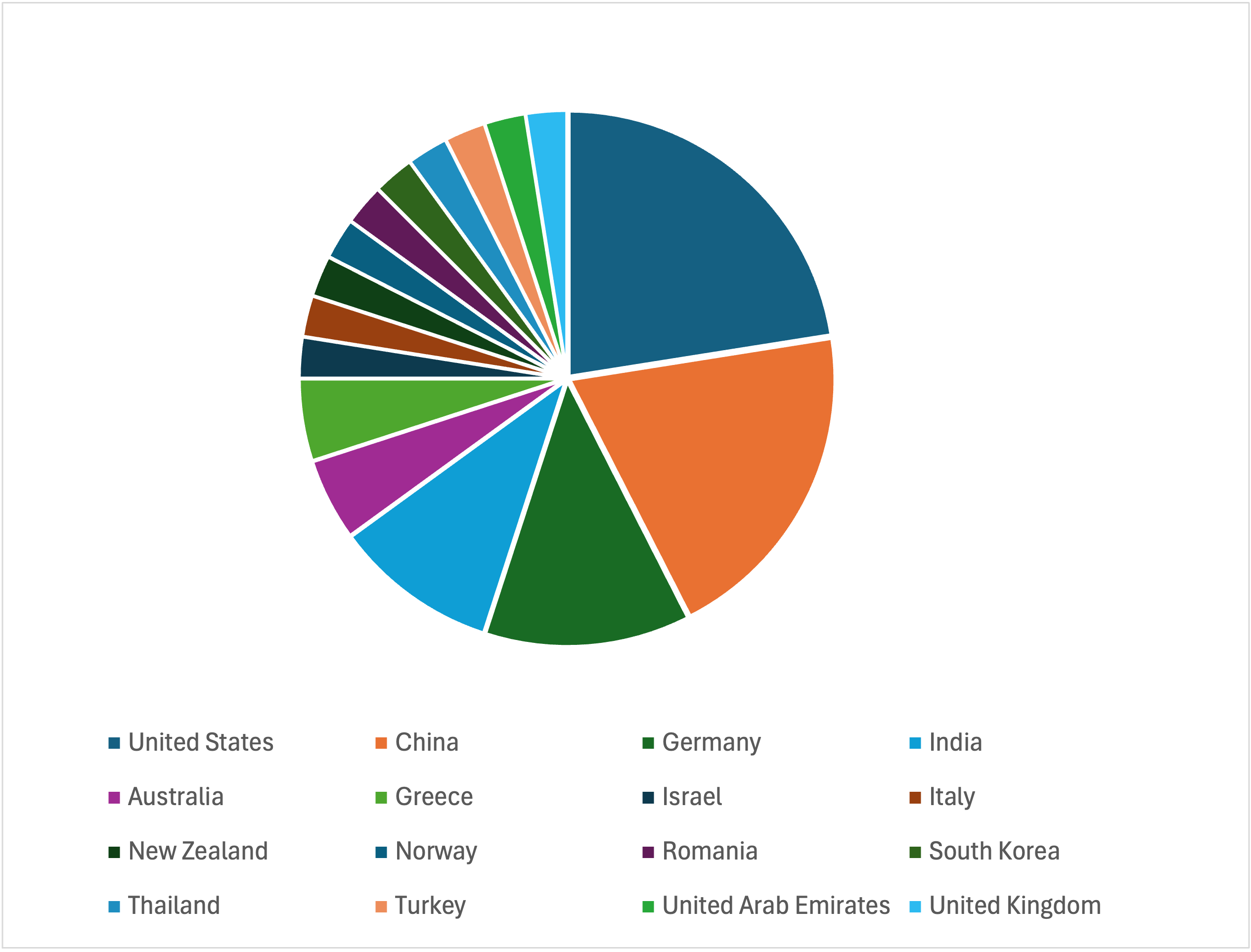}
\caption{Country-wise distribution of the 40 selected teams participating in the LiveRAG Challenge. %\oren{since we provide the number pf participating teams I would use a simple list of country-number of teams instead of a pie chart or maybe a map (we have one so no additional work is needed)}\david{+1, or to take it out, this chart is good for presentation but is not really needed in the report.}
} \label{fig:country-pie}
\end{figure}
}

\begin{table}[ht]
\caption{Distribution of accepted teams by country.}
\label{tab: accepted teams distribution}
\centering
\Comment{
\begin{table}[ht]
\centering
\begin{tabular}{rl|l}
\multicolumn{3}{c}{{\bf Team Distribution by Country}}\\
\toprule
Rank & Country & Team Count \\
\midrule
1 & United States & 9 \\
2 & China & 8 \\
3 & Germany & 5 \\
4 & India & 4 \\
5 & Australia & 2 \\
5 & Greece & 2 \\
7 & New Zealand & 1 \\
7 & Thailand & 1 \\
7 & Turkey & 1 \\
7 & Norway & 1 \\
7 & Italy & 1 \\
7 & United Arab Emirates & 1 \\
7 & South Korea & 1 \\
7 & United Kingdom & 1 \\
7 & Israel & 1 \\
7 & Romania & 1 \\
\bottomrule
\end{tabular}
\vspace{0.1cm}
\caption{Distribution of accepted teams by country.}
\label{tab:country-distribution}
\end{table}

\begin{table}[ht]
\centering
\begin{tabular}{r|l|l}
\multicolumn{3}{c}{{\bf Team Distribution by Country}}\\
\toprule
Rank & Team Count & Countries \\
\midrule
1 & 9 & United States \\
2 & 8 & China \\
3 & 5 & Germany \\
4 & 4 & India \\
5 & 2 & Australia, Greece \\
6 & 1 & Israel, Italy, New Zealand, Norway, Romania, South Korea, Thailand, Turkey,\\
  &   & United Arab Emirates, United Kingdom \\
\bottomrule
\end{tabular}
\vspace{0.1cm}
\caption{Distribution of accepted teams by country.}
\label{tab:country-distribution}
\end{table}
}

\begin{tabular}{lr}
%\multicolumn{2}{c}\\
\toprule
Countries & Team Count \\
\midrule
United States & 9 \\
\cmidrule{1-2}
China & 8 \\
\cmidrule{1-2}
Germany & 5 \\
\cmidrule{1-2}
India & 4 \\
\cmidrule{1-2}
Australia, Greece & 2 \\
\cmidrule{1-2}
Israel, Italy, New Zealand, Norway, Romania, South Korea, & 1 \\
Thailand, Turkey, United Arab Emirates, United Kingdom & \\
\midrule
Overall     & 40 \\
\bottomrule
\end{tabular}
\end{table}

\subsection{Technical Approaches}
%While we had very solid engagement from many teams during the development our kickoff and education sessions, 
As expected, not all teams made it to the LiveRAG challenge Day, primarily due to the demanding task of building a state-of-the-art RAG system within the limited time frame of the challenge. Nevertheless, we were pleased to observe that 25 teams were able to participate and answer 500 questions within a 2-hour time limit. Table \ref{tab:leaderboard} reports the list of active teams that submitted valid answers on time. 

Based on participant responses to an online survey, and their final reports~\citep{ran2025rmitliverag,
cofala2025ragtifierliverag,
salemi2025ciirliverag,
kim2025ltrrliverag,
ruangtanusak2025ped100xliverag,
martinez2025clueragliverag,
besrour2025ragentaliverag,
bakagianni2025topclustragliverag,
zhou2025knowledgeliverag,
dong2025leveragingliverag,
lajewska20225liverag,
duh2025hltcoeliverag}, 
we observed that most teams utilized \datamorgana{} for evaluation and training, and expressed their satisfaction with the tool. For query rewriting, most teams employed an LLM to decompose, rephrase, or expand the original question. Magikarp \citep{zhou2025knowledgeliverag} augmented the question with knowledge elements extracted from top-ranked retrieval results and used the expanded version for re-ranking. UDInfo~\citep{martinez2025clueragliverag} applied rule-based query re-writers - one tailored for sparse and one for dense retrieval.

For retrieval, most teams use a hybrid search over the two pre-built indices. A few teams built their own search index, rephrasing the documents using an LLM or using advanced embedders. Ped100x~\citep{ruangtanusak2025ped100xliverag} classified documents into a predefined topic taxonomy, which was then used to prune the search results based on the question topic. All teams re-ranked the search results using a variety of cross-embedding-based re-rankers such as BGE-m3, Jina-m0, and Cohere-3.5. RMIT-ADMS \citep{ran2025rmitliverag} applied an LLM-based re-ranker to assess whether a document contains the information needed to answer the question. The documents were then re-ranked according to the LLM's generated odds for the token “Yes”. 

For prompt generation, most teams augmented the question with 3-10 retrieved passages, with the exception of PRMAS-DRCA, %\citep{raj-mall2025prmas-drcaliverag} 
which used 50 passages. Unlike other teams, Ragtifier~\citep{cofala2025ragtifierliverag} added the passages to the prompt in reverse order of their retrieval score. BagBag\footnote{\url{https://huggingface.co/datasets/LiveRAG/Reports/resolve/main/SIGIR_2025_LiveRAG-Workshop_2617.pdf}} aggressively truncated and summarized the passages to reduce the size of the augmented prompt. All teams employed state-of-the-art LLMs (such as Claude-sonnet, Deepseek-R1, GPT-4o) for evaluation, comparing their generated answer with the one provided by \datamorgana{}.

%In summary, based on the survey, most active participants demonstrated high levels of engagement and high overall satisfaction with the challenge experience. All survey participants expressed a willingness to recommend the challenge to others and stated that they are interested in participating in future LiveRAG events.

\subsection{Live Challenge Day}

A “dry” test session with a small set of 50 questions was conducted a week prior to the actual live event to allow participants test their system and submission process. 

During the LiveRAG challenge Day, which took place on May 12, 2025, participants received a set of 500 synthetic questions questions, automatically generated by \datamorgana{}.  The participating teams were split into two sessions based on their preferred time-zone, each with their own benchmarks. A seed set of 105 shared questions was embedded within the two benchmarks, for manual evaluation (see \S{\ref{sec: Manual Evaluation}}), validation of the LLM-based judgment, and calibration across sessions.
For each question, participants were requested to provide a structured response consisting of
%\Ni
%\Nii
% \Niii
\begin{enumerate}[itemsep=-1mm]
    \item the answer generated by their RAG solution,
    \item the passages used for prompt augmentation, and
    \item the final prompt submitted to to Falcon3 for answer generation. 
\end{enumerate}
%\oren{please have a look}
%\david{rewritten -consider moving it to 2.2 (when Pinecone is described).}
Most teams participating in the Live event were using the prebuilt indices, and in particular the Pinecone index. To serve queries in a fault tolerant, low latency manner, Pinecone 
%clusters separate routing and query execution in the following way: every query is being routed by the Query Router to the relevant 
routes the query to a relevant
Query Executor
%that handles the Index - Slab
\cite{ingber2025accurate},   
%ensuring that even if many users  issue queries at the same time, Pinecone is able to distribute the load  
ensuring load balancing followed by a 
%so that each user experiences 
consistent, fast response time.
%In addition, Pinecone's Slab architecture ensures that regardless of the index size, query latency remains low, as the system dynamically compacts and re-indexes data.
Despite the heavy load during the Live Day Challenge sessions, we were pleased to see the smooth and effective service provided by Pinecone, OpenSearch, and Falcon3, enabling participants to answer the challenge questions on time.

\Comment{
\oren{this comes without a proper context. We should mention that most teams participating in the Live event were using the pre-built indices} To serve query results in a fault tolerant, low latency manner, Pinecone clusters separate routing and query execution in the following way: every query is being routed by the Query Router to the relevant Query Executor that handles the Index - Slab\oren{add \cite{ingber2025accurate}}. This ensures that even if many users  issue queries at the same time, Pinecone is able to distribute the load so that each user experiences consistent, low latency\oren{"query responses."} queries. In addition,\oren{"Pinecone's Slab architecture..."} the Slab architecture ensures that regardless of the index size, query latency remains low, as the system dynamically compacts and re-indexes data.}

To qualify for evaluation, competitors had to upload their results (following the requested format) to the Hugging Face LiveRAG space, within a strict two-hour time limit (an average time of $\sim$14 seconds per question). 
The participating teams were split into two sessions based on their preferred time-zone. 
%, where leaderboard scores were published afterwards.

\section{Evaluation}
\label{sec:eval}
The evaluation was conducted in two stages. First, all submissions were evaluated using LLM-as-a-judge~\citep{gu2024survey}. Then, the top-ranked submissions were manually evaluated by the organizers to validate the LLM-based assessments and to determine the final winners.

\subsection{LLM-based evaluation}

For LLM-based judgment we used Claude-3.5-sonnet\footnote{
\url{https://www.anthropic.com/news/claude-3-5-sonnet}
}, a state-of-the-art LLM, instructed to assess system-generated answers based on two metrics, Correctness and Faithfulness, defined below. 
Let $a$ be a system response to question $q$ with a corresponding reference answer $r$.

%\yoelle{please shorten the correctness and faithfulness paragraphs by at least by 50\%?}

\paragraph*{{\bf Correctness}} is measured by two components:
\begin{itemize}
\item {\bf Coverage}: The portion of vital information from the reference answer that is covered by the generated answer. This metric is highly inspired by the nugget-based metric proposed in TREC 2024 RAG track~\citep{pradeep2024initialnuggetevaluationresults},
%ym too much info on DataMorgana - removing a few sentences
and is computed as follows. We first instruct the LLM,
%using the prompt appearing in Appendix \S{\ref{app:prompts}}, Fig.~\ref{fig:prompt_statement_extraction},
to extract atomic claims from the reference answer. Then, each claim $c$ is classified as \texttt{Direct}, \texttt{Useful}, or \texttt{Useless}. %, or \texttt{Conversational}.
%, using the prompt appearing in Appendix \S{\ref{app:prompts}}, Fig.~\ref{fig:prompt_relatedness}. 
Then, we run an LLM-based \textit{Natural Language Inference} (NLI) 
%ym removing all prompt references
%(Appendix \S{\ref{app:prompts}}, Fig.~\ref{fig:prompt_nli}) 
to verify whether the generated answer implies the \texttt{Direct} and \texttt{Useful} claims. $NLI(a, c)$ is 1 if $a$ entails $c$, 0 if $a$ is neutral to $c$, and -1 if $a$ contradicts $c$. The coverage score is defined as: 
\begin{equation}
Cov(a,r)=\alpha \sum_{c \in D_r}\frac{NLI(a, c)}{|D_r|} + (1-\alpha)\sum_{c \in U_r}\frac{NLI(a, c)}{|U_r|}\ ,
\end{equation}
where $D_r$ and $U_r$ are the sets of \texttt{Direct} and \texttt{Useful} claims, respectively. Finally, $\alpha$ controls the weights assigned to the \texttt{Direct} and \texttt{Useful} terms of the formula\footnote{If $U_r$ is empty, the corresponding term is excluded from the formula and $\alpha$ is set to 1.}. We used $\alpha=0.7$ in our evaluation.
%the work in \citep{pradeep2025great}.
\item {\bf Relatedness:} The portion of vital claims in the generated answers that are related to the given question. Following the same procedure applied for the Coverage metric, we extract and classify claims from the generated answer $a$. 
%Depending on whether the question $q$ requires a concise answer or not\footnote{This condition depends on whether $q$ has been generated using a specific question category for answer conciseness.}, 
The relatedness score is defined as:
\begin{equation}
Rel(a) = \frac{|D_a|}{|D_a| + |I_a|}
\end{equation}
% \[
% Rel(a, q) = 
% \begin{cases}
% \frac{|D_a|}{|D_a| + |U_a| + |I_a|},& \text{if } q \text{requires a concise answer}\\
%  \\
%  \frac{|D_a|}{|D_a| + |I_a|}, & \text{otherwise}
% \end{cases}\ ,
% \]
where $D_a$ %, $U_a$, 
and $I_a$ are the sets of \texttt{Direct}, %\texttt{Useful}, 
and \texttt{Useless} claims appearing in $a$, respectively.

\end{itemize}
Finally, the harmonic mean of two metrics, graded on a continuous scale, is defined as the answer's Correctness score\footnote{We scaled the Correctness score to $[-1,..2]$, following the scoring mechanism suggested for the CRAG challenge \cite{yang2024crag}.}.
%\simo{can we report all scores in [0, 1] instead of [0,2] so that the statement about the harmonic mean holds?}. %, using the following aggregation formula:
%\dc{This will not work as we already published the leaderborads on [-1..2] scale.}

% \[
% Cor(q,a) = \begin{cases}
% Cov(a,r) & \text{if } Cov(a,r) \le 0 \\
% \\
% \frac{4 \cdot Cov(a,r) \cdot Rel(a, q)}{Cov(a,r) + Rel(a,q)} &\text{Otherwise}
% \end{cases}
% \]

% with the following representative points of the final Correctness metric:
% \begin{itemize}
% \item 2: Fully covered and fully related (no irrelevant information)
% \item 1: Fully covered, but contains additional irrelevant or redundant information
% \item 0: No answer provided (abstention, e.g., ``I don't know'')
% \item -1: Not covered and unrelated
%\end{itemize}

\paragraph*{{\bf Faithfulness}}
assesses whether the response is grounded in the retrieved passages. This metric revisits the metric provided by the Ragas system \citep{es2024ragas}. 
Specifically, given a question $q$, a generated answer $a$, and the set of retrieved documents $R$, we extract the claims appearing in the answer $a$, and assess whether each claim is entailed by at least one passage $r \in R$. Accordingly, the Faithfulness score is:

\begin{equation}
F(a, R) = \sum_{c \in C_a}\frac{NLI(c, R)}{|C_a|}\ ,
\end{equation}
where $C_a$ is the set of claims appearing in answer $a$, and $NLI(c,R)=\max_{r\in R} NLI(c,r)$.

% The Faithfulness score is graded on a continuous scale with the following representative points:
% \begin{itemize}
% \item 1: Full support. All answer claims are grounded
% \item 0: Neutral. The retrieved context $R$ does not support any claim  of the generated answer
% \item -1: Contradiction. All answer claims contradict the retrieved content $R$
% \end{itemize}

Both Correctness and Faithfulness contributed to the final evaluation score. Partial submissions were allowed, i.e., participants could skip some of the questions which were considered as abstentions. Due to evaluation budget constraints, while there was no limit on answer length, only the first 300 words of each response were evaluated. Furthermore, for Faithfulness computation, only the first 10 passages in the submitted list were considered.

\subsection{Manual Evaluation}\label{sec: Manual Evaluation}
In addition to the LLM-based evaluation process, the answers of the top-13 performing teams were manually evaluated by more than a dozen of qualified annotators, using the following metrics (all are on [0-2] Likert scale):
\begin{itemize}
\item {\bf Coverage:} how many of the vital claims, appearing in the reference answer, are covered by the generated answer.
\item {\bf Relatedness:} the portion of vital claims in the generated answer that relates to the question
% \begin{itemize}
% \item 0: Unrelated
%  \item 1: A non-negligible portion of the answer is related
%  \item 2: Fully related (all claims of the generated answer are related)
 % \end{itemize}
\item {\bf Quality:} Answer quality is subjectively evaluated while considering answer length, fluency, bad language, usage of unspecified terms or out-of-context terms, etc.
\end{itemize}
We aggregated the three metrics per question using the Borda counts~\citep{dwork2001rank}, and then averaged them across all questions to determine each team's final manual score.

\section{Results}
%ym moved the first sentence to the LiveChallenge Day section
% The participating teams were split into two sessions based on their preferred time-zone. During the Live Challenge Day (May 12), a benchmark of 500 questions was released to the teams shortly before each session. Participants were required to submit their answers within a two-hour time window. 

% A seed set of 105 shared questions was embedded within the two benchmarks, for manual evaluation (see \S{\ref{sec: Manual Evaluation}}), validation of the LLM-based judgment, and calibration across sessions. 
The leaderboards of the LLM-based scores for the two sessions are given in Table \ref{tab:leaderboard}, along with the results of Falcon3-10B-Instruct operating without RAG, serving as a na\"ive baseline.
\begin{table}
\scriptsize
\centering
\caption{\label{tab:leaderboard}LiveRAG Challenge - Leaderboards.}
\begin{tabular}{llllll}
\multicolumn{6}{c}{Session 1- May 12, 2025, 07:00 - 09:00 UTC}\\
\toprule
Rank   &Team ID &Team Name    &Organization          &Correctness[-1:2]  &Faithfulness[-1:1]\\
\toprule
1      &2615   &RMIT-ADMS      &RMIT, Australia    &{\bf 1.199317}           &{\bf 0.477382}\\
2      &2587   &RUC\_DeepSearch& Renmin University, China  &0.969273           &0.387808\\
3      &2620   &Ped100X        &SCBX, Thailand     &0.928893           &0.043381\\
4      &2677   &PRMAS-DRCA         &Indian Institute of Science    &0.922780       &0.410600\\
5      &2668   &Hybrid Search Graph  &Southwest University, China          &0.875091         &0.315802\\
6      &2617   &BagBag  &Hefei University, China            &0.694073           &-0.911353\\
7      &2669   &UniClustRAG        &University of Ioannina, Greece	 &0.685146           &0.460062\\
8      &2624   &METURAG            &Middle East Technical U., Turkey&0.673451           &0.325339\\
9      &2643   &DeepRAG            &New York University, UAE    &0.566053           &0.097828\\
10     &2635   &UiS-IAI            &University of Stavanger, Norway&0.552328           &0.433697\\
11     &2665   &SNU-LDILab         &Seoul National University, South Korea&0.517367           &0.103027\\
12     &2586   &Gravitational Lens &University of Auckland, New-Zeland &0.376637           &-0.988097\\
\midrule
Falcon3 (NO-RAG) &    &   &   &0.339 & ---\\
%Session 2 - May 12, 15:00 - 17:00 UTC
\bottomrule
\\
\\
\multicolumn{6}{c}{Session 2 - May 12, 2025, 15:00 - 17:00 UTC}\\
\toprule
Rank   &Team ID &Team Name    &Organization          &Correctness[-1:2]  &Faithfulness[-1:1]\\
\toprule
1      &2636   & Magikarp  &Chinese Academy of Sciences         &{\bf 1.231578}   &{\bf 0.656464}\\
2      &2596   & UDInfo    &University of Delaware, USA       & 1.200586 &0.623175\\
3      &2614   & RAGtifier &L3S Research Center, Germany         & 1.134454 &0.552365\\
4      &2626   & HLTCOE    &Johns Hopkins University, USA       & 1.070111 &0.340711\\
5      &2591   & Ragmatazz &Open Source Connections          & 1.011956 &0.519394\\
6      &2611   & ScaledRAG &UMASS, USA          & 0.996348 &0.418273\\
7      &2664   & Emorag   & Emory University, USA       & 0.890718 &0.556581\\
8      &2671   & Graph-Enhanced RAG&Huawei Technologies, UK           & 0.875714 &0.529335\\
9      &2650   & Multi-Agent Adaptive RAG& TU Dresden, Germany          & 0.836110 &0.200420\\
10     &2660   & Starlight     &CMU, USA      & 0.818337 &0.433003\\
11     &2648   & NoobRAG       &TU Dresden, Germany    & 0.655292 &0.154648\\
12      &2580   & UIUC-RAGents  &U. Illinois at Urbana Champaign, USA	         & 0.565043 &-0.302616\\
13      &2652   & AugmentRAG-TUD&Snowflake, US           & 0.532533 &0.655634\\
\midrule
Falcon3 (NO-RAG) &    &   &   &0.307 &---\\
\bottomrule
\end{tabular}
\end{table}

As can be seen in Table \ref{tab:leaderboard}, all teams performed better than Falcon3-10B-Instruct with no RAG support (referred to as ``Falcon3 (NO-RAG)'' in the Table), achieving a better Correctness score, and demonstrating RAG significant contribution for question answering. 
In terms of Faithfulness, most teams scored positively, i.e., their answers were inferred directly from the retrieved content used for augmentation. A few teams were scored negatively. One of these teams truncated aggressively the augmenting passages, a decision that significantly hurt their Faithfulness score.
%\david{check BagBag, Gravitational (no report), UIUC (a strange reranker & probably bugs in passage filtering) }

Furthermore, the answers of the 13 leading teams according to the LLM-based Correctness score (top-5 from Session 1 and top-8 from session 2), were manually evaluated by the organizers. The team answers for the shared seed set of 105 questions were evaluated for their the Coverage, Relatedness, Quality, and aggregated Borda score ((see \S{\ref{sec: Manual Evaluation}}).
The teams' manual scores, along with their LLM-based Correctness scores over the 105 questions in the shared set, are presented in Table \ref{tab:manual-lb}.
\begin{table}
\scriptsize
\centering
\caption{\label{tab:manual-lb}LiveRAG Challenge - Manual evaluation of 13 leading teams over the 105 shared questions. The LLM-based Correctness score over the 105 questions is given for reference.}
\begin{tabular}{lllllll|l}
\multicolumn{8}{c}{{\bf Manual Evaluation}}\\
\toprule
Rank & Team ID & Team Name & Borda & Coverage & Relatedness & Quality & LLM-Correctness \\
\midrule
1 & 2615 & RMIT-ADMS & \bf{7.706731} & \bf{1.615385} & \bf{1.884615} & \bf{1.673077} & 1.120858 \\
2 & 2614 & RAGtifier & 7.350962 & 1.557692 & 1.817308 & 1.557692 & {\bf 1.140128} \\
3 & 2596 & UDInfo & 7.240385 & 1.509615 & 1.855769 & 1.548077 & 1.109643 \\
4 & 2636 & Magikarp & 7.076923 & 1.451923 & 1.875000 & 1.567308 & 1.122471 \\
5 & 2620 & Ped100X & 6.225962 & 1.307692 & 1.692308 & 1.451923 & 0.787572 \\
6 & 2611 & ScaledRAG & 6.072115 & 1.221154 & 1.730769 & 1.442308 & 0.887048 \\
7 & 2626 & HLTCOE & 6.019231 & 1.278846 & 1.692308 & 1.403846 & 0.959300 \\
8 & 2591 & Ragmatazz & 5.471154 & 1.086538 & 1.769231 & 1.230769 & 0.892480 \\
9 & 2677 & PRMAS-DRCA & 5.206731 & 1.298077 & 1.480769 & 1.269231 & 0.801354 \\
10 & 2668 & Hybrid Search with Graph & 5.076923 & 1.173077 & 1.730769 & 1.125000 & 0.664194 \\
11 & 2587 & RUC\_DeepSearch & 5.067308 & 1.134615 & 1.653846 & 1.221154 & 0.702122 \\
12 & 2671 & Graph-Enhanced RAG & 4.802885 & 1.144231 & 1.576923 & 1.240385 & 0.777058 \\
13 & 2664 & Emorag & 4.682692 & 1.086538 & 1.326923 & 1.134615 & 0.854235 \\
\bottomrule
\end{tabular}
\end{table}

Interestingly, there is a high correlation between the LLM-based Correctness scores, as measured over the shared set of 105 questions, and the manual scores of the 13 leading teams. Table \ref{tab:correlation} depicts the Pearson correlation between these scores. The LLM-based score is mostly correlated with the aggregated Borda score, while Relatedness, for which most teams scored extremely high, is moderately correlated. 
Moreover, Table \ref{tab:manual-lb} clearly shows that the four leading teams, according to the LLM-based correctness score, also lead in terms of their manual scores.

\begin{table}
\scriptsize
\centering
\caption{\label{tab:correlation}Pearson correlation between the LLM-based Correctness scores of the 13 leading teams, over the shared set of 105 questions, and their manual evaluation scores.}
\begin{tabular}{lll}
\toprule
LLM-metric      & Manual metric     &Pearson \\
\toprule
\multirow{4}{*}{Correctness}     
     & Borda             & 0.8826\\
     & Coverage          & 0.8240\\
     & Quality           & 0.8490\\
     & Relatedness       & 0.6021\\
\bottomrule
\end{tabular}
\end{table}

\section{Summary}
The LiveRAG Challenge provided an opportunity for participating teams to develop real-time retrieval-augmented generation (RAG) systems for question answering, with a focus on retrieval, prompt generation, evaluation, and answer validation. The availability of free resources for selected participants, including pre-built indices, free access to Falcon3-10B-Instruct, and DataMorgana, significantly lowered the entry barrier, led to a high number of responses for the challenge's call for participation.

Out of the 40 selected teams, 25 actively participated in the live event and successfully submitted their answers within the allotted time frame. All participating teams outperformed the Falcon3-10B-Instruct baseline without RAG. Manual evaluation results were highly consistent with those from the LLM-based evaluation, supporting the robustness of our LLM-based evaluation methodology. It is worth mentioning that  DataMorgana was adopted by all teams for generating question–answer pairs
for training and evaluation.

Based on the received feedback, the challenge generated significant interest and enthusiasm. Many participants expressed interest in joining future LiveRAG Challenge events. We are considering organizing the challenge again next year, with potentially 
%higher question difficulty and 
extended variety of question types.

\paragraph*{Acknowledgments:}
We thank our sponsors Amazon AWS, Pinecone, and Hugging Face, who made this Challenge quite unique by providing credits to selected participants. Special thanks to Hakan Gokalp and Shlomi Shemesh at AWS and Michelle Habonneau and Thomas Wolff at Hugging Face for their generous support and help.
We are as always grateful to our colleagues and awesome partners at TII and AI71, in particular Hitanshu Shah, Dharansh Patel, Darshan Agarwal, Pranjal Dave, Ramy Makary, and Chaouki Kasmi. We thank our Program Committee members Charles L. A. Clarke, Yi Chang,  Ido Guy, Oren Kurland, Yiqun Liu, Antonio Mallia, Marc Najork,  Fabrizio Silvestri, Ian Soboroff,  Emine Yilmaz, and Elad Yom-Tov for assisting in reviewing all LiveRAG workshop reports. 
%\oren{we got assistant from other AI71 personnel. Let me check. I think we should also thank the SIGIR people for assisting us}
%\yoelle{who else helpded with \datamorgana{} service?} 

%\balance%
%\bibliographystyle{ACM-Reference-Format}
%\bibliographystyle{plain}
\bibliographystyle{unsrt}
\bibliography{biblio}

\begin{thebibliography}{10}

\bibitem{lewis2020retrieval}
Patrick Lewis, Ethan Perez, Aleksandra Piktus, Fabio Petroni, Vladimir Karpukhin, Naman Goyal, Heinrich K{\"u}ttler, Mike Lewis, Wen-tau Yih, Tim Rockt{\"a}schel, et~al.
\newblock Retrieval-augmented generation for knowledge-intensive {NLP} tasks.
\newblock {\em Advances in Neural Information Processing Systems}, 33:9459--9474, 2020.

\bibitem{izacard2022few}
Gautier Izacard, Patrick Lewis, Maria Lomeli, Lucas Hosseini, Fabio Petroni, Timo Schick, Jane Dwivedi-Yu, Armand Joulin, Sebastian Riedel, and Edouard Grave.
\newblock Atlas: Few-shot learning with retrieval augmented language models.
\newblock {\em Journal of Machine Learning Research}, 24(251):1--43, 2023.

\bibitem{Gao+al:24a}
Yunfan Gao, Yun Xiong, Xinyu Gao, Kangxiang Jia, Jinliu Pan, Yuxi Bi, Yi~Dai, Jiawei Sun, Meng Wang, and Haofen Wang.
\newblock {Retrieval-Augmented Generation for Large Language Models: A Survey}.
\newblock \url{https://arxiv.org/abs/2312.10997}, 2024.

\bibitem{pradeep2025great}
Ronak Pradeep, Nandan Thakur, Shivani Upadhyay, Daniel Campos, Nick Craswell, and Jimmy Lin.
\newblock The great nugget recall: Automating fact extraction and {RAG} evaluation with large language models.
\newblock {\em arXiv preprint arXiv:2504.15068}, 2025.

\bibitem{es2024ragas}
Shahul Es, Jithin James, Luis~Espinosa Anke, and Steven Schockaert.
\newblock Ragas: Automated evaluation of retrieval augmented generation.
\newblock In {\em Proceedings of the 18th Conference of the European Chapter of the Association for Computational Linguistics: System Demonstrations}, pages 150--158, 2024.

\bibitem{filice2025generatingdiverseqabenchmarks}
Simone Filice, Guy Horowitz, David Carmel, Zohar Karnin, Liane Lewin-Eytan, and Yoelle Maarek.
\newblock Generating diverse {QA} benchmarks for {RAG} evaluation with {DataMorgana}.
\newblock \url{https://arxiv.org/abs/2501.12789}, 2025.

\bibitem{yang2024crag}
Xiao Yang, Kai Sun, Hao Xin, Yushi Sun, Nikita Bhalla, Xiangsen Chen, Sajal Choudhary, Rongze Gui, Ziran Jiang, Ziyu Jiang, et~al.
\newblock {CRAG}-comprehensive {RAG} benchmark.
\newblock {\em Advances in Neural Information Processing Systems}, 37:10470--10490, 2024.

\bibitem{thakur2025support}
Nandan Thakur, Ronak Pradeep, Shivani Upadhyay, Daniel Campos, Nick Craswell, and Jimmy Lin.
\newblock Support evaluation for the trec 2024 {RAG} track: Comparing human versus llm judges.
\newblock {\em arXiv preprint arXiv:2504.15205}, 2025.

\bibitem{rahmani2025}
Hossein~A. Rahmani, Varsha Ramineni, Nick Craswell, Bhaskar Mitra, and Emine Yilmaz.
\newblock Towards understanding bias in synthetic data for evaluation, 2025.

\bibitem{soboroff2025don}
Ian Soboroff.
\newblock Don’t use llms to make relevance judgments.
\newblock {\em Information retrieval research journal}, 1(1):10--54195, 2025.

\bibitem{penedo2024the}
Guilherme Penedo, Hynek Kydl{\'\i}{\v{c}}ek, Loubna~Ben allal, Anton Lozhkov, Margaret Mitchell, Colin Raffel, Leandro~Von Werra, and Thomas Wolf.
\newblock The fineweb datasets: Decanting the web for the finest text data at scale.
\newblock In {\em The Thirty-eight Conference on Neural Information Processing Systems Datasets and Benchmarks Track}, 2024.

\bibitem{wang2024textembeddingsweaklysupervisedcontrastive}
Liang Wang, Nan Yang, Xiaolong Huang, Binxing Jiao, Linjun Yang, Daxin Jiang, Rangan Majumder, and Furu Wei.
\newblock Text embeddings by weakly-supervised contrastive pre-training, 2024.

\bibitem{ingber2025accurate}
Amir Ingber and Edo Liberty.
\newblock Accurate and efficient metadata filtering in pinecone's serverless vector database.
\newblock In {\em Proceedings of the 1st Workshop on Vector Databases (VecDB@ICML2025)}, Vancouver, Canada, July 2025.

\bibitem{ailon.2009}
Nir Ailon and Bernard Chazelle.
\newblock The fast johnson–lindenstrauss transform and approximate nearest neighbors.
\newblock {\em SIAM Journal on Computing}, 39(1):302--322, 2009.

\bibitem{jegou.2011}
Herve Jegou, Matthijs Douze, and Cordelia Schmid.
\newblock Product quantization for nearest neighbor search.
\newblock {\em IEEE Transactions on Pattern Analysis and Machine Intelligence}, 33(1):117--128, 2011.

\bibitem{malkov.2020}
Y.~A. C.Malkov and D.~A. Yashunin.
\newblock Efficient and robust approximate nearest neighbor search using hierarchical navigable small world graphs.
\newblock {\em IEEE Transactions on Pattern Analysis and Machine Intelligence}, 42(4):824--836, 2020.

\bibitem{filice2025dmacl}
Simone Filice, Guy Horowitz, David Carmel, Zohar Karnin, Liane Lewin-Eytan, and Yoelle Maarek.
\newblock Generating {Q\&A} benchmarks for {RAG} evaluation in enterprise settings.
\newblock In {\em Proceedings of the 63st Annual Meeting of the Association for Computational Linguistics (Industry Track)}, 2025.

\bibitem{ran2025rmitliverag}
Kun Ran, Shuoqi Sun, Khoi Nguyen~Dinh Anh, Damiano Spina, and Oleg Zendel.
\newblock {RMIT-ADM+S at the SIGIR 2025 LiveRAG Challenge}.
\newblock \url{https://arxiv.org/abs/2506.14516}, 2025.

\bibitem{cofala2025ragtifierliverag}
Tim Cofala, Oleh Astappiev, William Xion, and Hailay Teklehaymanot.
\newblock {RAGtifier: Evaluating RAG Generation Approaches of State-of-the-Art RAG Systems for the SIGIR LiveRAG Competition}.
\newblock \url{https://arxiv.org/abs/2506.14412}, 2025.

\bibitem{salemi2025ciirliverag}
Alireza Salemi, Mukta Maddipatla, and Hamed Zamani.
\newblock {CIIR@LiveRAG 2025: Optimizing Multi-Agent Retrieval Augmented Generation through Self-Training}.
\newblock \url{https://arxiv.org/abs/2506.10844}, 2025.

\bibitem{kim2025ltrrliverag}
To~Eun Kim and Fernando Diaz.
\newblock {LTRR: Learning To Rank Retrievers for LLMs}.
\newblock \url{https://arxiv.org/abs/2506.13743}, 2025.

\bibitem{ruangtanusak2025ped100xliverag}
Saksorn Ruangtanusak, Natthapath Rungseesiripak, Peerawat Rojratchadakorn, Monthol Charattrakool, and Natapong Nitarach.
\newblock {DoTA-RAG: Dynamic of Thought Aggregation RAG}.
\newblock \url{https://arxiv.org/abs/2506.12571}, 2025.

\bibitem{martinez2025clueragliverag}
Damian Martinez, Catalina Riano, and Hui Fang.
\newblock {PreQRAG -- Classify and Rewrite for Enhanced RAG}.
\newblock \url{https://arxiv.org/abs/2506.17493}, 2025.

\bibitem{besrour2025ragentaliverag}
Ines Besrour, Jingbo He, Tobias Schreieder, and Michael Färber.
\newblock {RAGentA: Multi-Agent Retrieval-Augmented Generation for Attributed Question Answering}.
\newblock \url{https://arxiv.org/abs/2506.16988}, 2025.

\bibitem{bakagianni2025topclustragliverag}
Juli Bakagianni, John Pavlopoulos, and Aristidis Likas.
\newblock {TopClustRAG at SIGIR 2025 LiveRAG Challenge}.
\newblock \url{https://arxiv.org/abs/2506.15246}, 2025.

\bibitem{zhou2025knowledgeliverag}
Tong Zhou.
\newblock Knowledge-aware diverse reranking for cross-source question answering.
\newblock \url{https://arxiv.org/abs/2506.20476}, 2025.

\bibitem{dong2025leveragingliverag}
Guanting Dong, Xiaoxi Li, Yuyao Zhang, and Mengjie Deng.
\newblock {Leveraging LLM-Assisted Query Understanding for Live Retrieval-Augmented Generation}.
\newblock \url{https://arxiv.org/abs/2506.21384}, 2025.

\bibitem{lajewska20225liverag}
Weronika Łajewska, Ivica Kostric, Gabriel Iturra-Bocaz, Mariam Arustashvili, and Krisztian Balog.
\newblock {UiS-IAI@LiveRAG: Retrieval-Augmented Information Nugget-Based Generation of Responses}.
\newblock \url{https://arxiv.org/abs/2506.22210}, 2025.

\bibitem{duh2025hltcoeliverag}
Kevin Duh, Eugene Yang, Orion Weller, Andrew Yates, and Dawn Lawrie.
\newblock {HLTCOE at LiveRAG: GPT-Researcher using ColBERT retrieval}.
\newblock \url{https://arxiv.org/abs/2506.22356}, 2025.

\bibitem{gu2024survey}
Jiawei Gu, Xuhui Jiang, Zhichao Shi, Hexiang Tan, Xuehao Zhai, Chengjin Xu, Wei Li, Yinghan Shen, Shengjie Ma, Honghao Liu, et~al.
\newblock {A survey on LLM-as-a-judge}.
\newblock {\em arXiv preprint arXiv:2411.15594}, 2024.

\bibitem{pradeep2024initialnuggetevaluationresults}
Ronak Pradeep, Nandan Thakur, Shivani Upadhyay, Daniel Campos, Nick Craswell, and Jimmy Lin.
\newblock Initial nugget evaluation results for the trec 2024 rag track with the autonuggetizer framework, 2024.

\bibitem{dwork2001rank}
Cynthia Dwork, Ravi Kumar, Moni Naor, and Dandapani Sivakumar.
\newblock Rank aggregation methods for the web.
\newblock In {\em Proceedings of the 10th international conference on World Wide Web}, pages 613--622, 2001.

\end{thebibliography}

\end{document}